\documentclass[10pt,twocolumn,letterpaper]{article}

\usepackage[applications]{wacv}
\usepackage{multirow}
\usepackage{makecell}
\usepackage{booktabs} 

\definecolor{wacvblue}{rgb}{0.21,0.49,0.74}
\usepackage[pagebackref,breaklinks,colorlinks,allcolors=wacvblue]{hyperref}
\usepackage{algorithm}
\usepackage{algorithmic}
\usepackage{makecell}

\usepackage{color}
\definecolor{julieta_colour}{RGB}{217,95,2} %

\def\wacvPaperID{*****} 
\def\confName{WACV}
\def\confYear{2026}

\title{Curriculum-Based Strategies for Efficient Cross-Domain Action Recognition}

\author{Emily Kim\\
Carnegie Mellon University\\
{\tt\small ekim2@andrew.cmu.edu}
\and
Allen Wu\\
Carnegie Mellon University\\
{\tt\small allenwu@andrew.cmu.edu}
\and
Jessica Hodgins\\
Carnegie Mellon University\\
{\tt\small jkh@andrew.cmu.edu}
}

\begin{document}
\maketitle
\begin{abstract}
Despite significant progress in human action recognition, generalizing to diverse viewpoints remains a challenge. Most existing datasets are captured from ground-level perspectives, and models trained on them often struggle to transfer to drastically different domains such as aerial views. This paper examines how curriculum-based training strategies can improve generalization to unseen real aerial-view data without using any real aerial data during training.

We explore curriculum learning for cross-view action recognition using two out-of-domain sources: synthetic aerial-view data and real ground-view data. Our results on the evaluation on order of training (fine-tuning on synthetic aerial data vs. real ground data) shows that fine-tuning on real ground data but differ in how they transition from synthetic to real. The first uses a two-stage curriculum with direct fine-tuning, while the second applies a progressive curriculum that expands the dataset in multiple stages before fine-tuning. We evaluate both methods on the REMAG dataset using SlowFast (CNN-based) and MViTv2 (Transformer-based) architectures.

Results show that combining the two out-of-domain datasets clearly outperforms training on a single domain, whether real ground-view or synthetic aerial-view. Both curriculum strategies match the top-1 accuracy of simple dataset combination while offering efficiency gains. With the two-step fine-tuning method, SlowFast achieves up to a 37\% reduction in iterations and MViTv2 up to a 30\% reduction compared to simple combination. The multi-step progressive approach further reduces iterations, by up to 9\% for SlowFast and 30\% for MViTv2, relative to the two-step method. These findings demonstrate that curriculum-based training can maintain comparable performance (top-1 accuracy within 3\% range) while improving training efficiency in cross-view action recognition.
\end{abstract} 
\section{Introduction}
\label{sec:intro}

\begin{figure}[!htbp]
\centering
\includegraphics[width=0.8\linewidth,trim=0 0 0 0, clip]{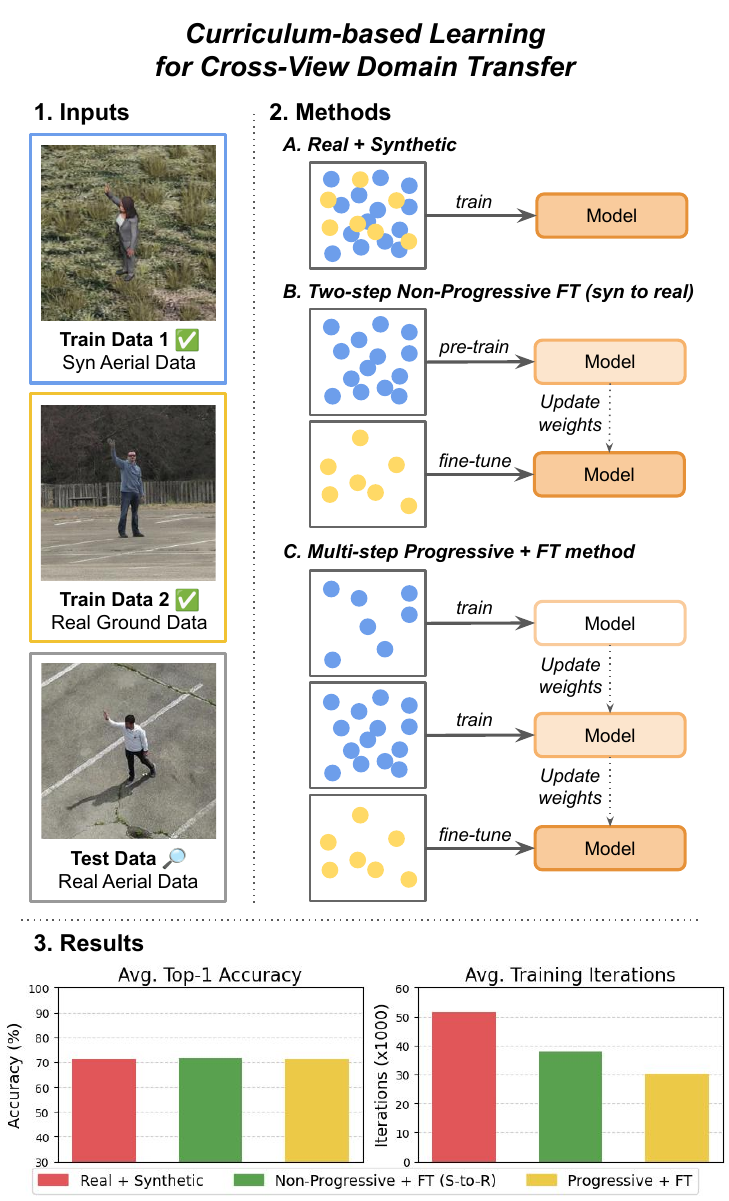}

\caption{\textbf{Curriculum-based learning strategies for cross-view domain transfer.} We compare three approaches for training action recognition models using two out-of-domain datasets: synthetic aerial-view and real ground-view videos. All models are evaluated on the unseen domain of real aerial-view videos. In this plot, we report the performance of the different methods on MViTv2 as the top-1 accuracy and the efficiency as the number of iterations (we fix all other parameters including batch size, input resolution, augmentations, and optimizer). Our results demonstrate that while all methods yield comparable top-1 accuracy, the progressive strategy achieves the highest training efficiency. 
}
\label{fig:teaser}
\end{figure}

While domain-specific datasets are highly effective for training action recognition models to perform well in a target test domain, they are not always readily available and may not have the required quantity or diversity of the data. This scarcity is particularly evident in aerial-view data, where collecting large-scale, domain-aligned video data is far more challenging than for standard ground-level views. As a result, progress in the field has often been driven by large-scale, general-purpose datasets such as Kinetics~\cite{kay2017kinetics}, which offer hundreds of action classes captured from standard ground-level viewpoints. However, models trained on such datasets often fail to generalize to novel viewpoints, particularly when transferring from ground-view to aerial-view scenarios. This performance degradation is largely due to the substantial domain shift introduced by the change in camera perspective~\cite{Panev_2024_WACV}.

To address this challenge, recent work has explored using synthetic data to augment or replace real-world video in low-resource or difficult-to-annotate settings. Datasets such as REMAG~\cite{Panev_2024_WACV}, RoCoG~\cite{deMelo2020vision}, and SURREACT~\cite{varol2021surreact} leverage computer-generated imagery or neural rendering to produce large-scale, multi-view action sequences with dense annotations. Among these, REMAG is notable for providing both real and synthetic data from ground and aerial viewpoints, enabling controlled studies of domain transfer. Studies on this dataset found that:
(i) synthetic data can significantly improve performance when paired with a small amount of domain-matched real data;
(ii) synthetic data alone remains insufficient, particularly for aerial views, due to appearance and motion realism gaps; and
(iii) training solely on ground-view data transfers poorly to aerial-view tasks.

Because aerial data collection requires substantial time, effort, and expertise, assembling large domain-specific datasets is often impractical. In such cases, synthetic datasets simulating aerial conditions offer a more accessible alternative. When real ground-view data for a matching domain is available, it becomes important to determine how synthetic aerial-view and real ground-view data can be effectively combined. These two sources are naturally complementary: synthetic aerial data provides viewpoint diversity, while real ground-view data contributes realistic appearance and motion cues. 

A straightforward approach is to combine the two datasets and train a model jointly. While this increases data diversity, it is unclear whether it is the most effective strategy. Prior work suggests that structuring the training process can improve both efficiency and performance~\cite{bengio2009curriculum}, rather than exposing the model to all samples at once. Curriculum learning addresses this by presenting data in a meaningful order, typically from easier to more difficult examples. Building on this idea, progressive learning incrementally expands the training set over time, offering further gains in training efficiency and stability~\cite{shen2023progressive}.

Motivated by these findings, we investigate curriculum-based and progressive training strategies for combining synthetic aerial-view and real ground-view data. Structured training schedules can help stabilize learning by introducing domain-shifted data in a controlled manner. Hence, in this paper, we assess whether such schedules can outperform naïve data combination in both performance and efficiency when generalizing to real aerial views.

In this work, we study curriculum-based learning for cross-view action recognition. Our training framework introduces synthetic aerial and real ground-view datasets in a structured manner to improve efficiency to real aerial-view scenarios. We train models using only synthetic aerial and real ground-view data, with evaluation performed exclusively on real aerial data. 

Our contributions are as follows:

\begin{itemize}
\item We investigate curriculum learning for cross-view action recognition, focusing on generalizing to real aerial-view data using synthetic aerial-view and real ground-view training data without access to any real aerial data during training.
\item We compare two curriculum strategies: a two-step fine-tuning method and a multi-step progressive learning method, each introducing domain-shifted data in structured ways to avoid potential confusion.
\item We conduct empirical evaluations on the REMAG dataset using two strong baseline models, SlowFast and MViTv2, reporting both top-1 accuracy and training efficiency.
\item Our results show that curriculum-based strategies achieve comparable performance while offering practical benefits in training efficiency compared to simply combining the two datasets, demonstrating their effectiveness as alternatives for domain-shifted action recognition.
\end{itemize}

\section{Related Work}
This section reviews three lines of work relevant to our method: (1) foundational models and datasets for video-based action recognition, (2) synthetic data for domain transfer, and (3) curriculum-based learning strategies for structured training with synthetic data.

\subsection{Core Action Recognition Models and Datasets}

Early deep-learning–based action recognition models, such as I3D~\cite{carreira2017quo}, extended 2D convolutional networks by inflating spatial kernels into the temporal domain, enabling direct modeling of spatiotemporal features. Building on this idea, SlowFast~\cite{feichtenhofer2019slowfast} introduced a dual-pathway design: a \emph{slow} pathway operating at a low frame rate to capture high-level semantic context, and a \emph{fast} pathway operating at a high frame rate to preserve fine-grained motion cues. This architecture improved temporal fidelity while maintaining computational efficiency, establishing SlowFast as a widely adopted benchmark model.

The field has since shifted toward Transformer-based architectures, which capture long-range dependencies and multiscale spatiotemporal patterns through self-attention. MViTv2~\cite{li2022mvitv2} enhances the original Multiscale Vision Transformer~\cite{fan2021multiscale} with pooling attention and improved positional encodings, yielding strong performance on large-scale video benchmarks. Video Swin Transformer~\cite{liu2022video_swin_transformer} adapts the Swin Transformer from images to video via shifted-window self-attention over spatiotemporal tokens, preserving locality-inductive biases while enabling efficient scaling. VideoMAE V2~\cite{wang2023videomaev2} advances masked autoencoding for video by introducing a dual-masking strategy—masking both encoder and decoder inputs—allowing efficient pre-training of billion-parameter video transformers and achieving state-of-the-art results across multiple video understanding tasks.

In this work, we select SlowFast and MViTv2 as our primary baselines, as they are widely recognized, well-studied architectures that represent complementary paradigms—CNN-based dual-pathway design and transformer-based multiscale attention—providing a balanced foundation for evaluating our domain-transfer methods.

These models are typically trained and evaluated on large-scale benchmarks such as Kinetics~\cite{kay2017kinetics}, which has expanded from 400 to over 700 action classes and contains more than 650{,}000 video clips. Earlier datasets, including UCF-101~\cite{soomro2012ucf101} and HMDB-51~\cite{kuehne2011hmdb}, remain popular for evaluation, particularly in studies on generalization and domain adaptation. In the aerial-view domain, UAV-Human~\cite{li2021uav} is among the largest benchmarks, comprising 155 action classes from 119 subjects with 22.5k video clips. Although smaller in scale, REMAG~\cite{Panev_2024_WACV} offers synchronized aerial and ground views alongside corresponding synthetic data, making it particularly well-suited for cross-view domain transfer experiments (Section~\ref{sec:domaintransfer}). The synchronized aerial–ground recordings in REMAG simplify cross-view learning by providing perfectly aligned temporal pairs, which reduce viewpoint ambiguity and enable cleaner supervision signals. However, collecting such data is challenging—requiring multi-camera setups with precise time synchronization, coordinated UAV and ground capture, and adherence to the logistical and regulatory constraints of aerial filming—and is generally impractical in real-world deployment scenarios where only a single viewpoint is available. This motivates our approach of using synthetic aerial-view data in combination with real ground-view data, enabling cross-view domain transfer without the need for synchronized multi-view recordings.

\subsection{Synthetic Data and Domain Transfer}\label{sec:domaintransfer}
Synthetic data has emerged as a scalable alternative to real video, particularly in domains where labeled data is scarce. Early work such as PHAV\cite{souza2016procedural} procedurally generated 39,982 synthetic clips using game engines to augment datasets like UCF-101 and HMDB-51. More recent pipelines, such as SURREACT~\cite{varol2021surreact}, employed Blender to animate 3D human models reconstructed from real motion using HMMR~\cite{humanMotionKanazawa19} and VIBE~\cite{kocabas2019vibe}, enabling synthetic-to-real generalization.

Other datasets, including RoCoG~\cite{deMelo2020vision} and RoCoG-v2~\cite{reddy2023synthetic}, targeted robot control tasks and provided paired synthetic and real videos for cross-domain benchmarking, highlighting synthetic data’s role in handling distribution shifts.

Recent studies have demonstrated that models trained entirely on synthetic data can be competitive with real-data baselines. Kim et al.\cite{kim2022how} introduced SynAPT, a unified suite of synthetic datasets, and showed that synthetic pretraining can outperform real-only training in low-resource regimes, even without overlapping label spaces. Guo et al.\cite{guo2022learning} similarly found that video representations learned solely from synthetic human motion can generalize effectively to real-world action recognition tasks without real supervision.

Panev et al.~\cite{Panev_2024_WACV} introduced REMAG, a comprehensive dataset containing real, CGI-rendered, and neural-rendered videos derived from both motion capture and monocular RGB tracking. Their benchmarks analyzed how different rendering methods and motion sources influence downstream performance. REMAG’s inclusion of synchronized aerial and ground views in both real and synthetic formats makes it a strong candidate for studying cross-view domain transfer. In our work, we directly leverage REMAG to evaluate performance across both domain and viewpoint shifts.

\subsection{Curriculum-based Learning with Synthetic Data}
Curriculum-based learning structures the training process by controlling the sequence and difficulty of training samples. Instead of exposing the model to all data simultaneously, training begins with simpler examples and gradually incorporates more complex ones. Bengio et al.\cite{bengio2009curriculum} introduced this concept, demonstrating improvements in convergence and generalization. Graves et al.\cite{graves2017automated} extended the approach with an automated sampling strategy based on a multi-armed bandit framework, dynamically selecting examples according to the model’s learning progress.

This strategy has been shown to benefit various vision tasks. For instance, Zhang et al.~\cite{zhang2017curriculum} applied a curriculum-based domain adaptation method to semantic segmentation, improving generalization from synthetic to real-world urban scenes.

\begin{figure*}[!htbp]
\centering
\includegraphics[width=\linewidth]{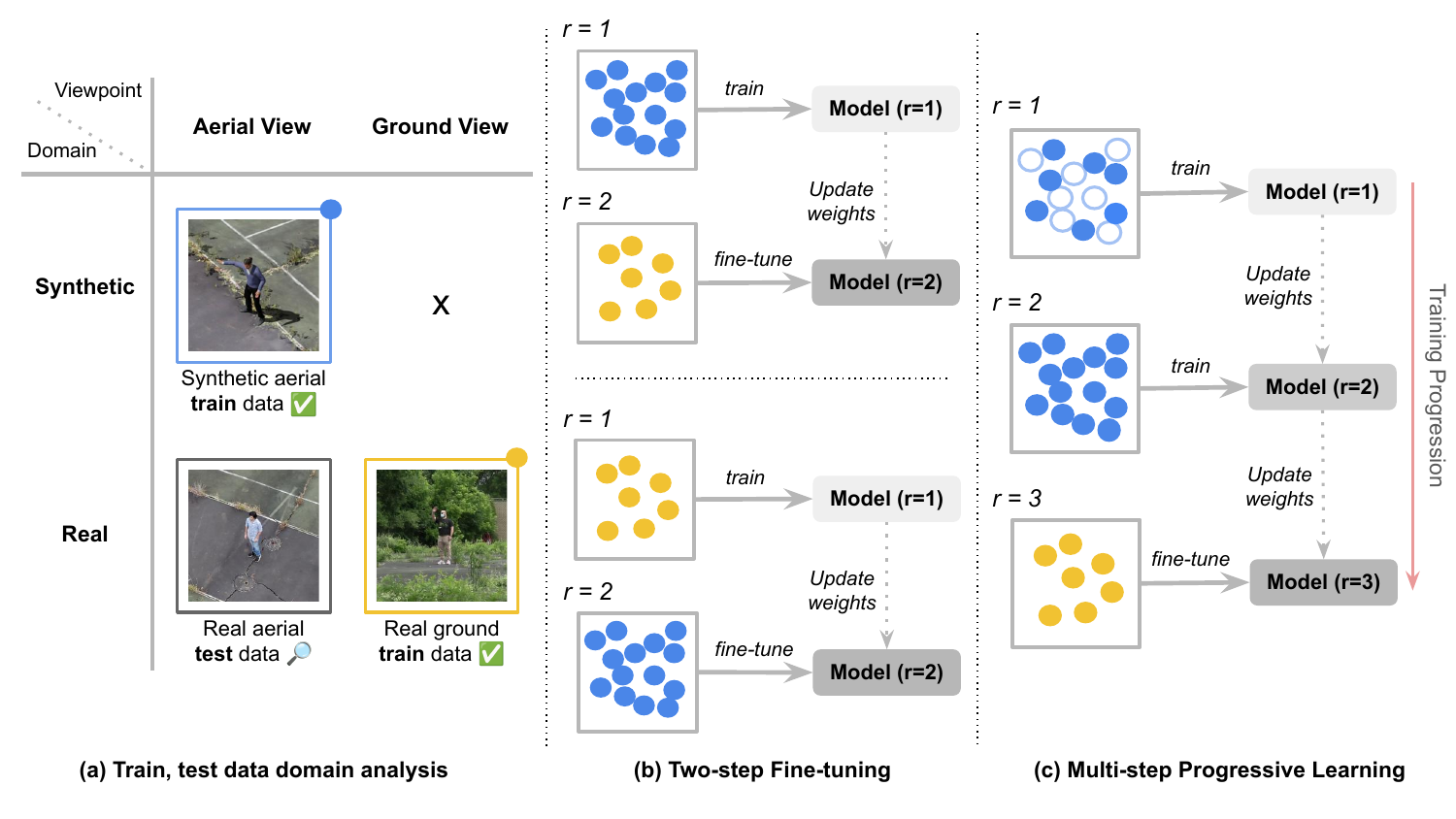}
\caption{\textbf{Overview of our progressive learning pipeline.} (a) Viewpoint–domain matrix of training and testing modalities. We use synthetic aerial and real ground-view data for training, and evaluate only on real aerial-view data. (b) Progressive learning procedure with $R=3$ rounds of training. The model is updated iteratively across rounds. The first $R{-}1$ rounds are trained with synthetic aerial data. In the final round, we compare two variants: (i) combined training using both synthetic aerial and real ground-view data, and (ii) fine-tuning using only real ground-view data.}
\label{fig:pipeline}
\end{figure*}

Recent work has explored progressive curriculum strategies, in which the size or difficulty of the training set increases over time. PDE~\cite{deng2023robust} adopted this approach to improve worst-group accuracy by 2.8\% while reducing training time by a factor of ten on classification benchmarks. PTL~\cite{shen2023progressive} applied progressive learning to object detection in UAV imagery, showing that gradually introducing synthetic data enhances cross-domain generalization.

Motivated by these results, we extend progressive curriculum strategies to cross-domain video-based action recognition. We specifically evaluate whether sequentially introducing synthetic and real data, either via multi-stage fine-tuning or progressive dataset expansion, can match or surpass static training with a combined dataset, while also improving training efficiency.

\section{Method}
\label{sec:method}

Our goal is to train a video-based action recognition model that generalizes well to real aerial data, without using any real aerial data during training. We compare three different training strategies using two out-of-domain datasets: synthetic aerial-view data ($\mathcal{D}_{\text{syn}}$) and real ground-view data ($\mathcal{D}_{\text{real}}$). The objective is to train efficiently and achieve high accuracy.

\paragraph{Naive combination (baseline).}  
The baseline strategy simply combines $\mathcal{D}_{\text{syn}}$ and $\mathcal{D}_{\text{real}}$ to train the model $\mathcal{M}_0$ in a single step. This approach is the most straightforward, given that both datasets are slightly shifted from the target domain.

\paragraph{Two-step fine-tuning.}  
The first curriculum-based learning method is a two-step fine-tuning strategy. We first pre-train the model $\mathcal{M}_0$ on one dataset and then fine-tune it on the other. In our main setting, we pre-train on $\mathcal{D}_{\text{syn}}$ and fine-tune on $\mathcal{D}_{\text{real}}$ to form $\mathcal{M}_2$. This sequential adaptation enables the model to gradually transition between domains. While the reverse direction (real-to-synthetic) is possible, we did not evaluate it here due to the smaller size and diversity of $\mathcal{D}_{\text{real}}$.

\begin{table*}[!htbp]
\centering

\begin{tabular}{ccc|cc}
\toprule
\multirow{2}{*}{\textbf{Training Strategy}} & \multicolumn{4}{c}{\textbf{Top-1 Accuracy (\%)}} \\
\cmidrule(lr){2-5}
                           & SlowFast (T1) & SlowFast (T2) & MViTv2 (T1) & MViTv2 (T2)\\
\midrule
Real Only (Aerial) & 63.25 & 68.36 & 72.43 & 72.90\\
\midrule
Real Only (Ground)                          & 41.58 & 43.53& 59.72 &61.96\\
Synthetic Only  (Aerial)                    & 52.07 & 53.46& 59.01 &64.31\\
\midrule
Real (G) + Synthetic (A) & 58.12& 59.01& \textbf{71.74}&71.02\\
Non-Progressive + FT (R-to-S)& 59.98 & 58.88& 69.26 & 67.69\\
Non-Progressive + FT (S-to-R)& \textbf{60.90}& \textbf{61.45}& \underline{70.82}& \textbf{72.48}\\
Progressive + FT& \underline{60.53}& \underline{59.56}&70.79&72.05\\
\bottomrule
\end{tabular}
\caption{\textbf{Top-1 Accuracy (\%) on Real Aerial Test Set across Training Strategies.}
We report results for SlowFast and MViTv2 models under different training strategies. T1 and T2 denote two distinct real aerial-view test sets. All models are trained on synthetic aerial-view and real ground-view data, and evaluated on unseen real aerial-view videos.}
\label{tab:strategy_acc}
\end{table*}

\paragraph{Multi-step progressive learning.}  
The second curriculum-based strategy is a multi-step progressive learning method. We train the model in three steps, where the first two steps are progressive expansion and the last step is the fine-tuning. In the first round, we train on a 50\% random subset of $\mathcal{D}_{\text{syn}}$, class-balanced and disjoint from the remaining data. In the second round, we expand training to the full $\mathcal{D}_{\text{syn}}$, allowing the model to adapt to increased domain complexity. Finally, we fine-tune the resulting model on $\mathcal{D}_{\text{real}}$. This gradual expansion allows the model to learn initial representations from a simpler synthetic domain before being exposed to the full data distribution. By structuring the learning process in this way, we avoid abrupt domain shifts and large optimization jumps, leading to more stable convergence and improved generalization. We did not explore progressive pretraining in the reverse direction due to the smaller scale of the real dataset. We also limited the number of rounds to three to make the experiments simple. 

We compare the performance and training efficiency of these three strategies to evaluate their ability to transfer to the real aerial domain. Our results suggest that curriculum-based strategies offer a robust alternative to naive domain mixing, improving both convergence stability and cross-view generalization.


\section{Experiments}

\begin{figure*}[!htbp]
    \centering
    \includegraphics[width=\linewidth]{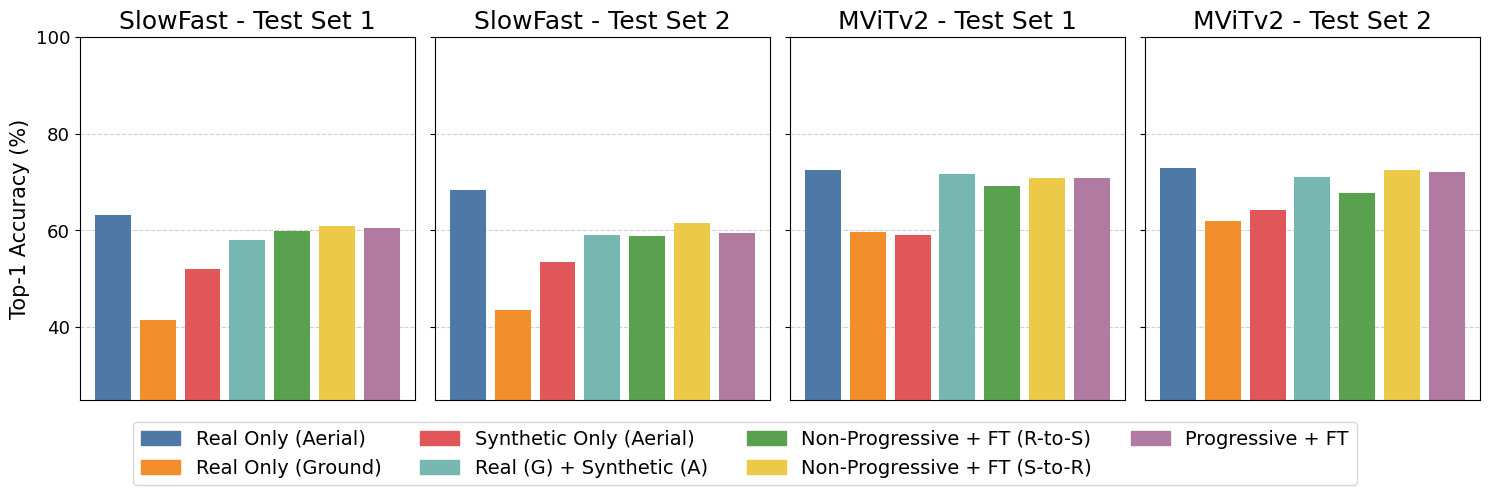}
    \caption{\textbf{Top-1 accuracy across training strategies for two test sets and two model architectures (SlowFast and MViTv2).} Each group corresponds to a training method: Real Only, Synthetic Only, Real (G) + Synthetic (A), Non-Progressive + FT (R-to-S), Non-Progressive + FT (S-to-R), and Progressive + FT. Bars are grouped by model type within each test set. Results show that combining real and synthetic data—especially through structured curricula—significantly improves generalization to the target aerial-view domain.}
    \label{fig:chart}
\end{figure*}

We evaluated our proposed progressive training approach on the REMAG~\cite{Panev_2024_WACV} dataset, which includes 11 action classes plus an idle class. For evaluation, we selected two distinct 5-subject sets (approximately 8k samples) from the real aerial-view data as test sets. For each test set, the corresponding training set consisted of real ground-view data covering 21 subjects, along with synthetic aerial-view data generated using CGI rendering and marker-based motion capture. Test subjects were excluded from all training data. 

To ensure balanced training across all classes, we applied oversampling such that each class contained approximately 6,000 samples in both the synthetic and real domains. Video clips were sampled at 64-frame intervals (30 fps, $\sim$2 seconds), and only clips in which at least 80\% of the frames shared the same class label were retained to ensure temporal consistency. For both SlowFast and MViTv2, we extracted 16 frames per clip, uniformly sampled every 4 frames within each 64-frame segment. We did not add any augmentation to the training data.

We evaluated our method using two state-of-the-art action recognition models: SlowFast~\cite{feichtenhofer2019slowfast} and MViTv2~\cite{fan2021multiscale}, both initialized with Kinetics-pretrained weights. Training was conducted on two NVIDIA GeForce RTX 4090 GPUs (each with 24 GB of memory). The batch size per GPU was 45 for SlowFast and 12 for MViTv2. For evaluation, we report top-1 classification accuracy on the real aerial-view test set.

\subsection{Experimental setup}
We compared the performance of models trained using five different strategies involving synthetic aerial-view and real ground-view data. First, we evaluated models trained using only synthetic aerial-view data and only real ground-view data, which serve as lower-bound baselines. We then evaluated three strategies described in Section~\ref{sec:method}: (1) naive combination of the two datasets, (2) two-step fine-tuning, and (3) multi-step progressive learning.

Progressive learning was conducted over $R = 3$ rounds. For each round $r$, we trained for $e_r$ epochs. For SlowFast, we used $e_1 = e_2 = 150$ epochs; for MViTv2, we used $e_1 = 30$ and $e_2 = 60$ epochs, based on our observation from the loss curve, making sure that the model was sufficiently trained on the provided training data without overfitting. In both models, the final round was trained until convergence. All models were optimized using AdamW. Learning rates were set to 0.1 for SlowFast and 1e-4 for MViTv2 during standard training. For fine-tuning, we reduced the learning rates to 0.05 for SlowFast and 5e-5 for MViTv2.

We ran the same series of experiments for two held-out test sets. We report top-1 accuracy for all experiments in Table~\ref{tab:strategy_acc} and in Figure~\ref{fig:chart}. We trained domain-specific models, each trained on \textit{Real Only (Aerial)}, \textit{Real Only (Ground)}, and \textit{Synthetic Only (Aerial)}. We label the naïve combination as \textit{Real (G) + Synthetic (A)}, the two-step fine-tuning method pretrained on real and fine-tuned on synthetic as \textit{Non-Progressive + FT (R-to-S)}, the two-step fine-tuning method pretrained on synthetic and fine-tuned on real as \textit{Non-Progressive + FT (S-to-R)}, and the multi-step progressive method as \textit{Progressive + FT}.

\begin{figure*}[!htbp]
\centering
\includegraphics[width=0.9\linewidth]{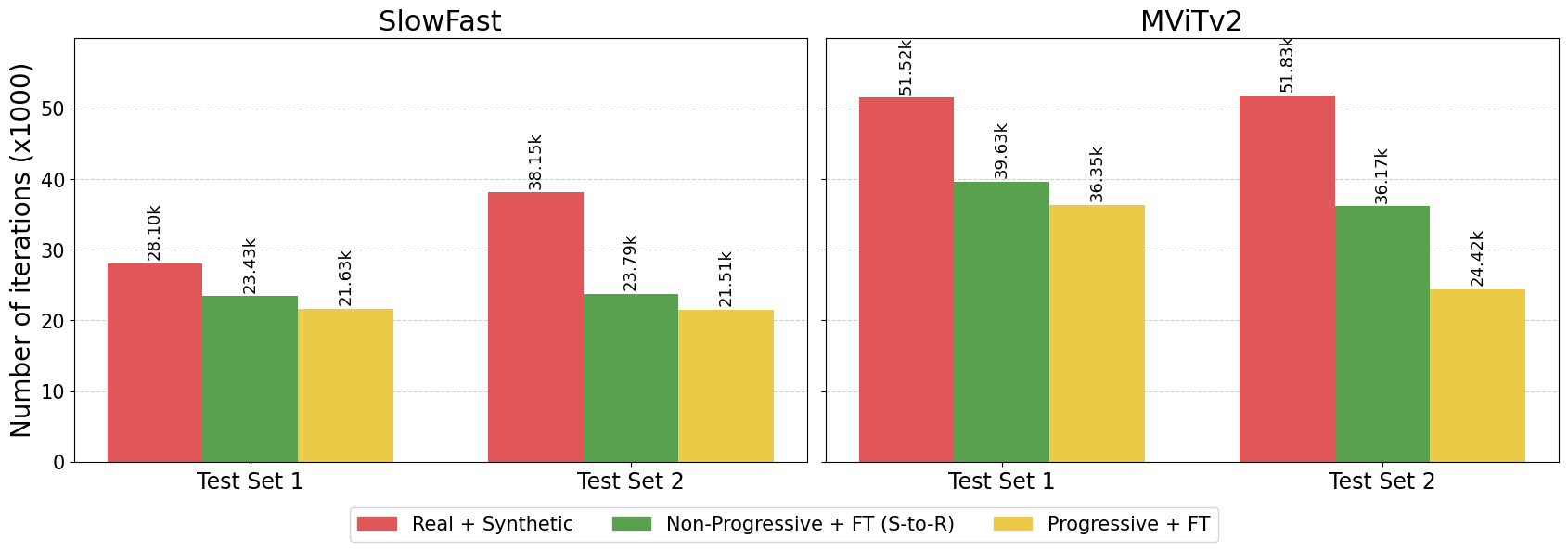}
\caption{\textbf{Evaluation of computational efficiency across training strategies.} We compare the total number of training iterations required for each method—\textit{Real (G) + Synthetic (A)}, \textit{Non-Progressive + FT (S-to-R)}, and \textit{Progressive + FT}—across two test sets for both SlowFast and MViTv2. While all methods achieve comparable top-1 accuracy, the\textit{ Progressive + FT }approach consistently requires fewer iterations, highlighting its efficiency. Notably, \textit{Progressive + FT} reduces the training cost by up to 30\% compared to naive or two-step fine-tuning strategies, making it a favorable option in resource-constrained settings. Since the batch size, input resolution, augmentations, and optimizer are fixed, the per-iteration cost is effectively constant, making iteration count a reliable proxy for wall-clock training time within a model.
}
\label{fig:efficiency}
\end{figure*}

In addition to evaluating model performance, we analyzed the computational efficiency of each method for combining the two training datasets. Because we used fixed batch sizes for all experiments, we measured efficiency by comparing the total number of training iterations required to complete training. Specifically, we evaluated: (1) naive combination of synthetic and real data, (2) two-step fine-tuning (synthetic to real), and (3) progressive curriculum learning using incrementally larger subsets of synthetic data followed by real data. The number of training iterations for each setup is shown in Figure~\ref{fig:efficiency}.

\subsection{Analysis of Training Strategy Components}
We now analyze the results in Figure~\ref{fig:chart}, Table~\ref{tab:strategy_acc}, and Figure~\ref{fig:efficiency}, comparing the different combinations of training methods in the following paragraphs: domain-specific training results, curriculum vs. non-curriculum results, training direction, and progressive vs. non-progressive scheduling.

\paragraph{Domain-Specific Training Results.}
We first trained the models on real aerial-view training data from the same domain as the test sets. SlowFast achieved 63\% and 68\% accuracy on Test Sets 1 and 2, respectively, while MViTv2 reached around 72\% on both. These results represent performance without domain shift and serve as an upper bound for the subsequent experiments.

In contrast, models trained solely on real ground-view or synthetic aerial-view data performed substantially worse. For SlowFast, the \textit{Real-only (Ground)} model achieved about 42\% top-1 accuracy, while the \textit{Synthetic-only (Aerial)} model reached roughly 53\%. MViTv2, by comparison, yielded similar performance in both cases, with top-1 accuracy around 60\%. These findings suggest that the transformer-based MViTv2 generalizes more robustly across domains, whereas the CNN-based SlowFast is more sensitive to domain shift. Notably, synthetic data can outperform real data when its viewpoint coverage better matches the target domain, as observed in SlowFast’s higher accuracy under synthetic-only training. Still, neither dataset alone provides strong generalization, underscoring the importance of combining complementary data sources.

\paragraph{Non-Curriculum-Based vs. Two-Step Fine-Tuning.}
Given the two distinct training sources, a natural baseline is to combine the datasets directly or to adopt a two-step fine-tuning approach. In terms of generalization performance, merging real and synthetic data (\textit{Real (G) + Synthetic (A)}) provides a clear improvement over single-domain training, achieving 59\% accuracy for SlowFast and 71\% for MViTv2. This demonstrates the benefit of increasing data diversity and domain coverage.

The two-step fine-tuning approach (\textit{Non-Progressive + FT}) achieves comparable or slightly better performance. For SlowFast, the \textit{S-to-R} variant consistently outperforms \textit{Real (G) + Synthetic (A)} across both test sets. For MViTv2, the difference is more nuanced: \textit{Real (G) + Synthetic (A)} performs slightly better on Test Set 1, while \textit{S-to-R} outperforms it on Test Set 2. These differences are modest (0.9–2.8\% for SlowFast and 0.9–4.8\% for MViTv2), but they suggest that a structured training sequence can provide benefits, particularly for architectures more sensitive to domain gaps like CNNs.

Where two-step fine-tuning clearly excels is in training efficiency. Across both models, it requires substantially fewer iterations to reach comparable accuracy. For SlowFast, \textit{S-to-R} reduces the iteration count by 6.5k (23\%) on Test Set 1 and 14.4k (37\%) on Test Set 2. For MViTv2, the reductions range from 11.9k (23\%) to 15.6k (30\%). Because the batch size and hardware are fixed, these reductions translate directly into proportional savings in compute time, resulting in faster training and lower resource usage without compromising accuracy.

\paragraph{Curriculum Direction.}
We examined whether the order of training affects generalization performance when using the two out-of-domain datasets: synthetic aerial ($\mathcal{D}{\text{syn}}$) and real ground-view ($\mathcal{D}{\text{real}}$). Each introduces a different domain gap: $\mathcal{D}{\text{syn}}$ lacks realism, while $\mathcal{D}{\text{real}}$ differs in viewpoint. It is unclear a priori whether training should begin with synthetic and then fine-tune on real (\textit{S-to-R}), or the reverse (\textit{R-to-S}).

As shown in Table~\ref{tab:strategy_acc}, \textit{S-to-R} consistently outperforms \textit{R-to-S} across both models and test sets. For SlowFast, \textit{R-to-S} achieves around 58–59\%, while \textit{S-to-R} reaches about 61\%. For MViTv2, \textit{R-to-S} yields 68–69\%, and \textit{S-to-R} improves to 71–72\%. While margins are small, the consistent trend suggests that starting with viewpoint-aligned (synthetic) data, even if unrealistic, supports better generalization than beginning with realistic but viewpoint-misaligned data. 
We hypothesize several factors contributing to this result:
\begin{itemize}
    \item \textbf{Dataset scale and diversity} – The larger synthetic dataset exposes the model to a broader range of spatiotemporal variations, enabling it to learn generalizable features before adapting to the real domain.
    \item \textbf{Clean labels and controlled conditions} – Synthetic data typically provides perfectly accurate labels and uniform coverage of rare poses, viewpoints, and lighting, reducing early training noise and improving representation quality.
    \item \textbf{Optimization stability} – Beginning with abundant, consistent synthetic data reduces the risk of early overfitting to the smaller real dataset, leading to more stable convergence and better final performance.
\end{itemize}

\paragraph{Progressive vs. Non-Progressive Scheduling.}
We compared the two curriculum-based strategies: (1) two-step fine-tuning (\textit{Non-Progressive + FT (S-to-R)}), and (2) progressive curriculum learning (\textit{Progressive + FT}). While the former is simpler, the latter systematically increases training diversity and can improve efficiency. In both cases, training begins with $\mathcal{D}_{\text{syn}}$ and transitions to $\mathcal{D}_{\text{real}}$.

In terms of performance, both strategies are comparable. For SlowFast, \textit{S-to-R} achieves around 61\%, while \textit{Progressive} yields 59.5–60.5\%. For MViTv2, the difference is less than 1\% across test sets.

However, the progressive method shows notable efficiency gains. For SlowFast, training iterations are reduced by 1.8k (7\%) on Test Set 1 and 2.3k (9\%) on Test Set 2. For MViTv2, reductions are 3.3k (8\%) and 11.8k (30\%), respectively. These results suggest that while both methods are similarly effective, the progressive approach achieves comparable performance with significantly less training effort.
\section{Discussion and Conclusion}

Our evaluation demonstrates that curriculum-based training strategies achieve performance comparable to naïvely training on the combined out-of-domain datasets (i.e., real ground-view and synthetic aerial-view videos), but with significantly greater training efficiency. Among the curriculum strategies, the multi-step progressive method was the most efficient, requiring substantially fewer training iterations to reach convergence.

Between the two non-progressive fine-tuning strategies, pre-training on viewpoint-aligned but less realistic synthetic aerial data followed by fine-tuning on real ground-view data (\textit{S-to-R}) consistently resulted in stronger generalization to the real aerial-view test domain. This finding underscores the importance of training order: aligning structural factors such as viewpoint early in the training process appears to provide a more stable foundation before introducing domain realism.

\subsection{Model-Level Observations}
We observed that model performance and sensitivity to training strategies varied across architectures. MViTv2, a transformer-based model, consistently outperformed the CNN-based SlowFast across all training regimes, including single-domain setups (e.g., real-only and synthetic-only). Moreover, MViTv2 exhibited greater robustness to different training curricula, showing only minor performance variation between strategies, while SlowFast showed larger gains from curriculum-based approaches. These findings are consistent with prior work suggesting that transformers, with their global receptive fields and reduced inductive biases, are more resilient to domain shifts. Notably, MViTv2 models trained with both real ground-view and synthetic aerial-view data achieved performance comparable to the no-domain-shift setting, further underscoring their robustness and the effectiveness of combining complementary data sources.

\subsection{Limitations}
Despite encouraging results, our study has several limitations. First, we evaluated only two model architectures (SlowFast and MViTv2), which limits the generalizability of our findings. Second, our progressive curriculum used a fixed three-step schedule for simplicity; adaptive or learnable curriculum schedules may further improve performance and efficiency. Third, we assumed access to clean, fully annotated data collected in matched capture environments for training and testing, where backgrounds, lighting, and camera setups were consistent across splits. While this isolates the effect of viewpoint change, it likely overestimates performance compared to real-world scenarios with mismatched conditions. Fourth, the dataset used is relatively small and constrained, potentially simplifying the domain adaptation problem. Finally, while our synthetic data offers strong viewpoint coverage, it lacks photorealism and scene complexity, which may reduce its effectiveness in generalizing to more realistic aerial environments.

\subsection{Future Work}
Future work should aim to expand the diversity of training data—adding more action classes, backgrounds, lighting conditions, and subject appearances—to better reflect real-world variability. Improving the realism of synthetic data, for example through appearance transfer or style adaptation, could help close the domain gap. Developing adaptive curriculum strategies that dynamically adjust the training schedule based on model feedback is another promising direction. Finally, evaluating curriculum-based methods across a wider range of architectures and datasets would help strengthen the external validity of our conclusions and facilitate broader adoption in practical deployment scenarios.

{
    \small
    \bibliographystyle{ieeenat_fullname}
    \bibliography{main}
}

\clearpage
\setcounter{page}{1}
\maketitlesupplementary

\section{Results: Confusion Matrices}

\begin{figure*}[tbhp!]
    \centering
    \includegraphics[width=0.6\linewidth]{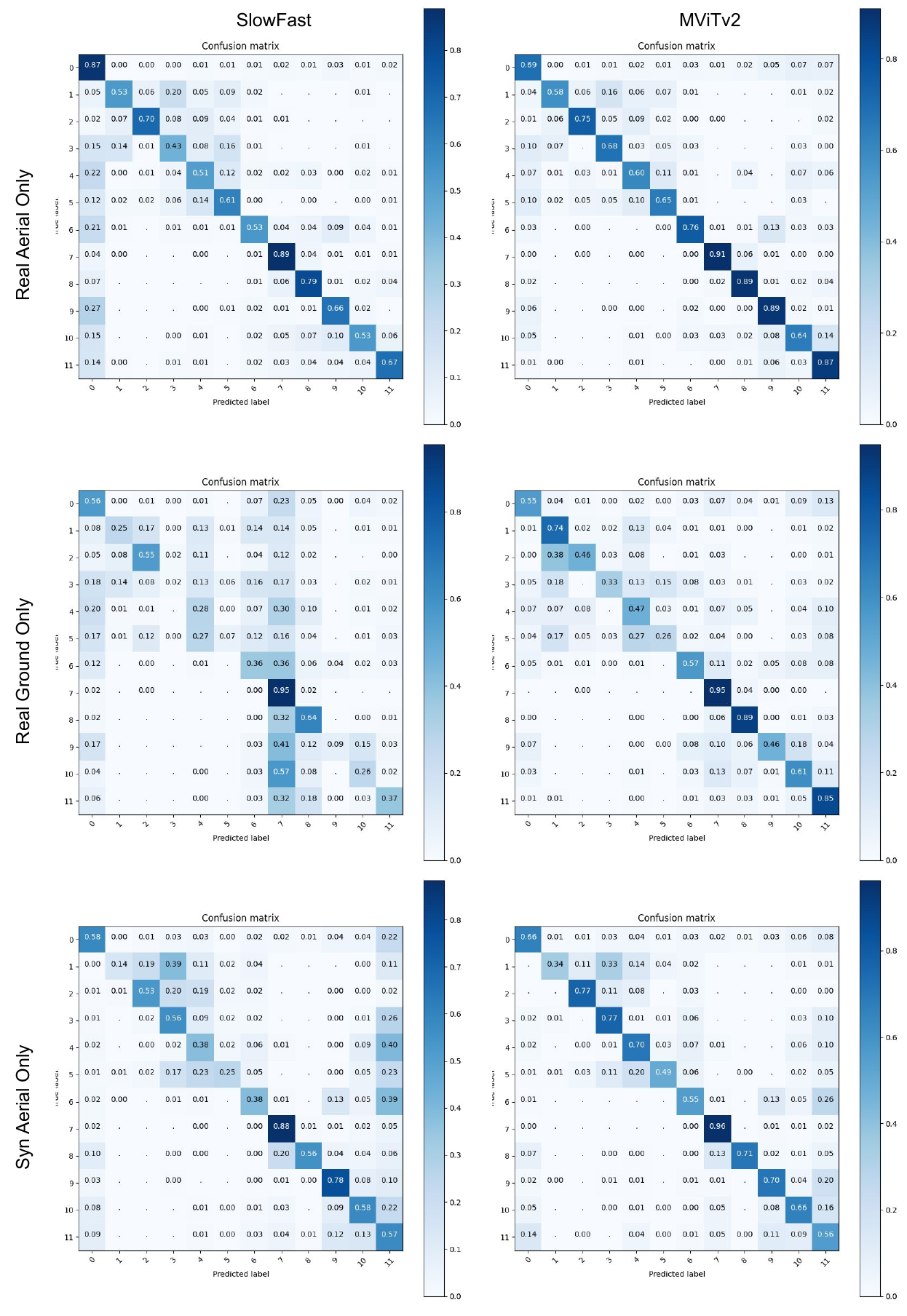}
    \caption{\textbf{Confusion Matrix 1} -- We show the detailed results for SlowFast and MViTv2. We observe the results for Real Aerial Only, Real Ground Only, and Synthetic Aerial Only.}
    \label{fig:conf1}
\end{figure*}

\begin{figure*}[tbhp!]
    \centering
    \includegraphics[width=0.6\linewidth]{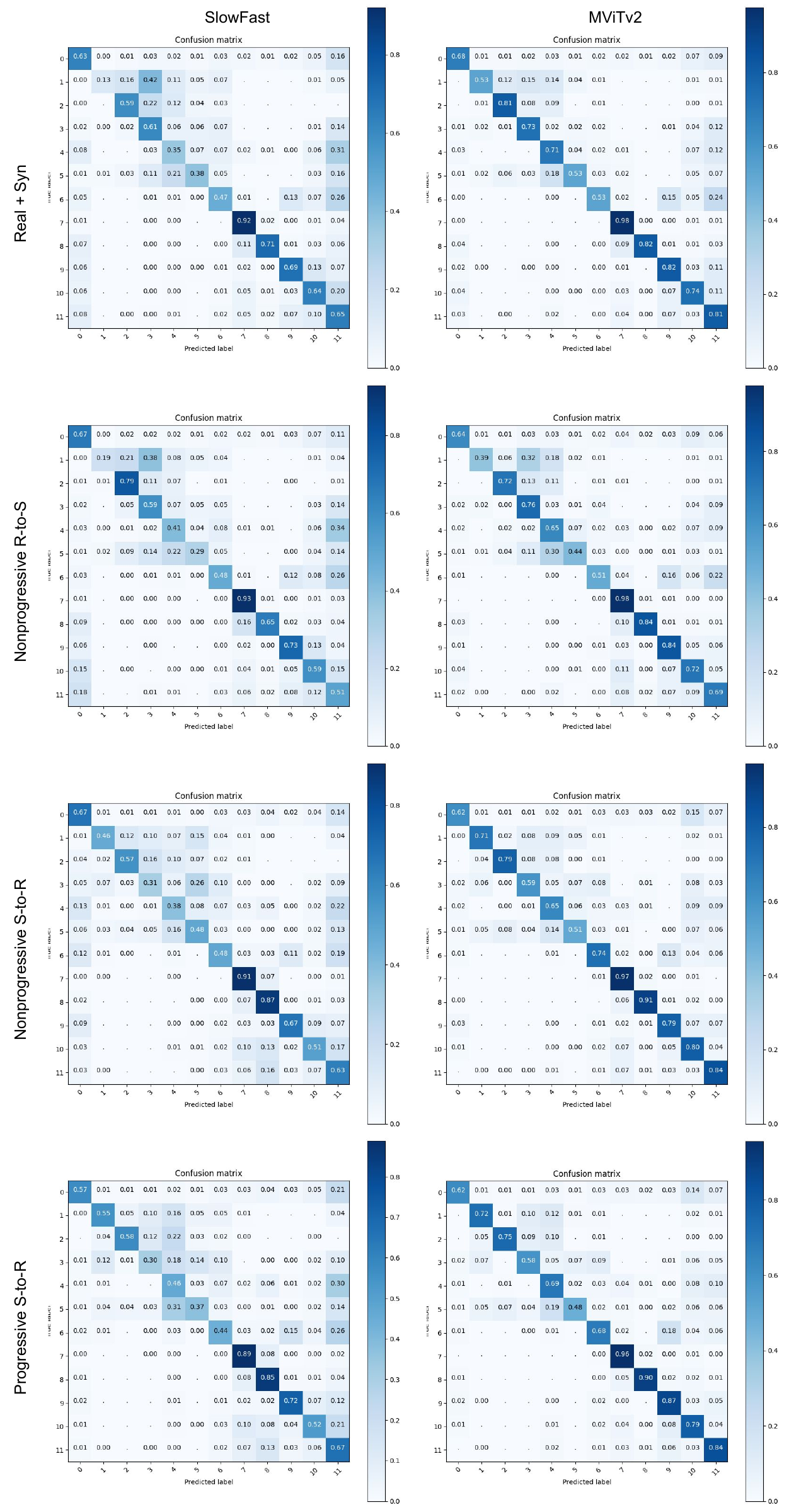}
    \caption{\textbf{Confusion Matrix 2} -- We show the detailed results for SlowFast and MViTv2. We observe the results for Real + Syn, Non-Progressive R-to-S, Non-Progressive S-to-R, and Progressive S-to-R.}
    \label{fig:conf2}
\end{figure*}

We report the confusion matrices from our experiments in Figures~\ref{fig:conf1} and~\ref{fig:conf2}. To avoid repetition, we report the matrices from Test Set 2 for both SlowFast and MViTv2. The actions are labeled as:
\begin{itemize}
\item \textbf{0:} ``idle"
\item \textbf{1:} ``wave"
\item \textbf{2:} ``wave attention"
\item \textbf{3:} ``shake fist"
\item \textbf{4:} ``move forward"
\item \textbf{5:} ``come here"
\item \textbf{6:} ``take picture with phone"
\item \textbf{7:} ``carry a shovel"
\item \textbf{8:} ``carry a bat"
\item \textbf{9:} ``carry a phone"
\item \textbf{10:} ``talk on a phone"
\item \textbf{11:} ``hold flashlight"
\end{itemize}

In Figure~\ref{fig:conf1}, we observe that SlowFast trained on real ground data struggles with classifying most actions, except ``idle," ``wave attention," ``carry a shovel," and ``hold a flashlight." The model trained on synthetic aerial data also struggles with actions that do not involve objects. MViTv2 does not confuse action classes as much as SlowFast, demonstrating the robustness of the transformer-based method, though its confusion matrices still show lighter colors when trained only on real ground data or only on synthetic aerial data.

When the two datasets are combined, the SlowFast model confuses ``wave" with ``shake fist," a similar pattern to the model trained using the \textit{Non-Progressive R-to-S} method. Overall, the performance of SlowFast improves significantly on object-involving actions when trained with both real and synthetic data. With MViTv2, we observe similar patterns in progressive and non-progressive methods pretrained on synthetic data and fine-tuned with real data, where the strongest intensity is seen along the diagonal of the confusion matrix. Overall, MViTv2 shows stronger intensity across the diagonal compared to the SlowFast, supporting the claim that the transformer-based model is more robust to unseen, real aerial-view data domain.

\end{document}